\documentclass[sigconf,authorversion,nonacm]{acmart}
\AtBeginDocument{%
  }

\setcopyright{acmlicensed}
\copyrightyear{2018}
\acmYear{2018}
\acmDOI{XXXXXXX.XXXXXXX}
\acmConference[Conference acronym 'XX]{Make sure to enter the correct
  conference title from your rights confirmation email}{June 03--05,
  2018}{Woodstock, NY}
\acmISBN{978-1-4503-XXXX-X/2018/06}

\usepackage{multirow}
\usepackage{listings} 
\usepackage{xcolor}   

\begin{document}

\title[A CVE-based Software Security Benchmark for LLMs]{TOSSS: a CVE-based Software Security Benchmark\\ for Large Language Models}


\author{Marc Damie}
\orcid{0000-0002-9484-4460}
\affiliation{%
	\institution{University of Twente}
	\city{Enschede}
	\country{The Netherlands}
}

\author{Murat Bilgehan Ertan}
\orcid{0009-0005-2651-7301}
\affiliation{%
	\institution{CWI Amsterdam}
	\city{Amsterdam}
	\country{The Netherlands}
}

\author{Domenico Essoussi}
\affiliation{%
	\institution{Erasmus University Rotterdam}
	\city{Rotterdam}
	\country{The Netherlands}
}

\author{Angela Makhanu}
\affiliation{%
	\institution{Datadog}
	\city{Paris}
	\country{France}
}

\author{Gaëtan Peter}
\affiliation{%
	\institution{Ecole Supérieure d'Ingénieurs Léonard de Vinci}
	\city{Courbevoie}
	\country{France}
}

\author{Roos Wensveen}
\affiliation{%
	\institution{Modat}
	\city{The Hague}
	\country{The Netherlands}
}

\renewcommand{\shortauthors}{Damie et al.}


\begin{abstract}
	With their increasing capabilities, Large Language Models (LLMs) are now used across many industries.
	They have become useful tools for software engineers and support a wide range of development tasks.
	As LLMs are increasingly used in software development workflows, a critical question arises: are LLMs good at software security?
	At the same time, organizations worldwide invest heavily in cybersecurity to reduce exposure to disruptive attacks.
	The integration of LLMs into software engineering workflows may introduce new vulnerabilities and weaken existing security efforts.

	We introduce TOSSS (Two-Option Secure Snippet Selection), a benchmark that measures the ability of LLMs to choose between secure and vulnerable code snippets.
	Existing security benchmarks for LLMs cover only a limited range of vulnerabilities.
	In contrast, TOSSS relies on the CVE database and provides an extensible framework that can integrate newly disclosed vulnerabilities over time.
	Our benchmark gives each model a security score between 0 and 1 based on its behavior; a score of 1 indicates that the model always selects the secure snippet, while a score of 0 indicates that it always selects the vulnerable one.
	We evaluate 14 widely used open-source and closed-source models on C/C++ and Java code and observe scores ranging from 0.48 to 0.89.
	LLM providers already publish many benchmark scores for their models, and TOSSS could become a complementary security-focused score to include in these reports.
\end{abstract}



\maketitle

\section{Introduction}
Recent advances in natural language processing~\cite{vaswani_attention_2017} have enabled the emergence of Large Language Models (LLMs) such as GPT \cite{openai_gpt5_system_card_2025}, LLaMA \cite{touvron_llama_2023}, and Gemini \cite{gemini_team_2024}, that can perform many tasks that were previously difficult to automate accurately and efficiently.
Modern LLMs can generate text whose quality is comparable to human-written documents in several domains.
Because of these capabilities, LLM-based services have been rapidly adopted across industries as well as by individual users.
This professional adoption is especially visible in software engineering~\cite{hou_large_2024}.
Companies have even designed services and models optimized for development tasks such as GitHub Copilot~\cite{mastropaolo_robustness_2023} or Mistral Codestral.

Unfortunately, this rapid adoption by software engineers is not without risks.
Soon after the introduction of coding assistants, several studies~\cite{khoury_how_2023,pearce_asleep_2025} reported security vulnerabilities in code generated by LLMs.
For example, Pearce et al.~\cite{pearce_asleep_2025} found that, among 1,689 programs generated with GitHub Copilot, 40\% contained security vulnerabilities.

These findings motivated subsequent work on benchmarking the software security skills of LLMs and identifying models that produce more secure programs
\cite{shahid_llm-csec_2025,li_safegenbench_2025,dora_hidden_2025,siddiq_sallm_2024,ryden_llmseccode_2024,dai_comprehensive_2025,wang_realsec-bench_2026,shen_secrepobench_2025,wang_is_2024,hajipour_codelmsec_2024,lian_se_2025,bhatt_purple_2023,chen_secureagentbench_2025,velasco_how_2025}.
These benchmarks compare LLMs across a range of programming tasks.
For each task, they typically rely on static analyzers to detect vulnerabilities in the generated code.

Although these benchmarks provide useful insights, they share a key limitation: \textbf{limited extensibility}.
Because they depend on fixed task sets, they cannot easily incorporate newly discovered vulnerabilities or additional programming languages because their coverage is constrained by the detection capabilities of the underlying static analyzers.
Furthermore, some approaches require \textbf{manual validation}, which introduces scalability limits and potential evaluation bias.

As cybersecurity evolves rapidly, an extensible benchmark is necessary to maintain an accurate security assessment of LLMs.
New vulnerabilities and weaknesses are continuously discovered, so benchmarks must be able to incorporate them over time.
Existing benchmarks depend on static analyzers, which restrict coverage to the detection capabilities and rule sets of these tools.
As a result, they cannot easily integrate newly disclosed vulnerabilities.
They also provide limited support for new programming languages.

To sum up, there is a need for an LLM security benchmark that can easily incorporate newly discovered vulnerabilities.

\paragraph{Our contributions}
We introduce a new security benchmark for LLMs called TOSSS (Two-Option Secure Snippet Selection).
To ensure extensibility, we adopt a strategy that differs from prior benchmarks.
Instead of asking LLMs to solve programming tasks, we present two versions of the same function, one vulnerable and one secure, and ask the model to select one.
The security metric is defined as the proportion of secure snippets selected by the model.
This metric is easy to interpret in practice.
A score close to 1 indicates consistent selection of secure code.
A score of 0.5 indicates performance comparable to random choice.
A score below 0.5 indicates a bias towards vulnerable code.

Unlike task-based benchmarks, TOSSS relies on pairs of functions composed of a vulnerable version and its corresponding secure version.
To construct these pairs, we build upon prior work in software repository mining, in particular MegaVul~\cite{ni_megavul_2024}.
MegaVul provides an automated pipeline to extract code associated with vulnerabilities reported in the CVE (Common Vulnerabilities and Exposures) database.
It enables retrieval of source code versions before and after a security fix.
By integrating this pipeline, our benchmark can continuously incorporate newly reported vulnerabilities and therefore remains extensible over time.

We execute our benchmark on 14 widely used open-source and closed-source models using both C/C++ and Java code snippets.
The resulting scores range from 0.48 to 0.89, highlighting substantial variation in secure coding behavior across models.
To support adoption by practitioners and encourage reporting of security scores, we release in open-source the TOSSS benchmark:

\url{https://github.com/MarcT0K/TOSSS-LLM-Benchmark}. 

\section{Related work}
\label{sec:related-work}
\begin{table*}
	\centering
	\begin{tabular}{|c|c|c|c|}
		\hline
		Benchmark                                          & LLM Task                      & Security Evaluation    & Test Case Generation               \\\hline\hline
		CyberSecEval~\cite{bhatt_purple_2023}              & Code Completion               & Static Analyzer        & Mining of Open-Source Repositories \\\hline
		SALLM~\cite{siddiq_sallm_2024}                     & Code Generation               & Static Analyzer        & Stack Overflow Mining              \\\hline
		LLMSecCode~\cite{ryden_llmseccode_2024}            & Code Generation + Code Repair & Static Analyzer        & Existing Research Dataset          \\\hline
		Wang et al.~\cite{wang_is_2024}                    & Code Generation + Code Repair & Custom Automated Tests & Manual Curation                    \\\hline
		CodeLMSec~\cite{hajipour_codelmsec_2024}           & Code Generation               & Static Analyzer        & LLM-Based                          \\\hline
		Dai et al.~\cite{dai_comprehensive_2025}           & Code Generation               & Static Analyzer        & Filtered Research Dataset          \\\hline
		SecRepoBench~\cite{shen_secrepobench_2025}         & Code Completion               & Fuzzer                 & Filtered Research Dataset          \\\hline
		ASE~\cite{lian_se_2025}                            & Code Completion               & Static Analyzer        & Mining of Open-Source Repositories \\\hline
		SecureAgentBench~\cite{chen_secureagentbench_2025} & Code Repair                   & Static Analyzer        & OSSFuzz Open-Source Repositories   \\\hline
		SafeGenBench~\cite{li_safegenbench_2025}           & Code Generation               & LLM + Static Analyzer  & LLM + Manual Curation              \\\hline
		Dora et al.~\cite{dora_hidden_2025}                & Code Generation               & Manual                 & Manual Curation                    \\\hline
		CWEval~\cite{peng_cweval_2025}                     & Code Generation               & Custom Automated Tests & Manual Curation                    \\\hline
		LLM-CSEC~\cite{shahid_llm-csec_2025}               & Code Generation               & Static Analyzer        & Manual Curation                    \\\hline
		SecCodeBench~\cite{chen_seccodebench-v2_2026}      & Code Generation               & LLM                    & Manual Curation                    \\\hline
		RealSec-Bench~\cite{wang_realsec-bench_2026}       & Code Generation               & Static Analyzer        & Mining of Open-Source Repositories \\\hline
		\hline
		TOSSS (\textbf{Ours})                              & \textbf{Code Selection}       & \textbf{Ground Truth}  & \textbf{CVE Mining}                \\\hline
	\end{tabular}
	\caption{Comparison of methodologies used in secure coding benchmarks for LLMs.}
	\label{tab:comp-sota}
\end{table*}

Following early observations of insecure coding practices in LLM outputs~\cite{khoury_how_2023}, several benchmarks were proposed to quantify this issue and compare available models.
Table~\ref{tab:comp-sota} summarizes the methodology of 15 existing secure coding benchmarks.
In this subsection, we discuss the main design choices observed in these works.

\paragraph{LLM task}
Evaluating the software security skills of LLMs is challenging because LLMs are used in different software engineering settings.
Developers may use LLMs to generate code from scratch, complete existing code, or repair vulnerable implementations.
Benchmark designers must therefore define a standardized task for evaluation.

As shown in Table~\ref{tab:comp-sota}, most benchmarks focus on code generation tasks, where models are instructed to produce code from scratch.
Some benchmarks also consider alternative settings, such as code repair~\cite{ryden_llmseccode_2024,wang_is_2024,chen_secureagentbench_2025} or code completion~\cite{bhatt_purple_2023,shen_secrepobench_2025,lian_se_2025}.

\paragraph{Security evaluation}
Beyond the task definition, the most critical design choice in a benchmark is the methodology used to assess the software security skills.
Most existing benchmarks rely on static analyzers such as CodeQL to detect vulnerabilities.
Although this approach is straightforward, it restricts the benchmark to the vulnerability classes supported by the selected analyzer.
As a result, extending coverage to newly disclosed vulnerabilities or additional programming languages may require significant effort.

A few benchmarks explore alternative evaluation strategies, including fuzzing~\cite{shen_secrepobench_2025}, LLM-based judging~\cite{li_safegenbench_2025,chen_seccodebench-v2_2026}, custom unit tests~\cite{wang_is_2024,peng_cweval_2025}, and manual verification~\cite{dora_hidden_2025}.
Fuzzing can uncover certain classes of vulnerabilities, but its coverage remains limited to specific execution behaviors.
Using an LLM as a security judge is problematic, since the objective of the benchmark is to evaluate the security reliability of LLMs.
Custom unit tests enable fine-grained analysis, but they require substantial expert effort to design and maintain.
Manual verification can provide accurate assessments, yet it does not scale and may introduce evaluator bias.

From this review, we identify two main limitations in existing methodologies.
First, they do not support straightforward extensibility.
Incorporating new vulnerabilities or programming languages typically requires modifying the underlying analyzer, fuzzer, or test suite.
Second, these approaches do not rely on explicit ground truth comparisons.
In `traditional' machine learning, evaluation is performed by comparing model outputs to labeled ground truth data.
In contrast, code generation tasks produce complex artifacts that cannot easily be matched to a predefined reference implementation.
Existing benchmarks therefore rely on external tools that approximate security assessment.

Our work introduces a novel benchmarking approach that enables security evaluation against explicit ground truth pairs, providing a consistent and robust assessment framework.

\paragraph{Test case generation}
Benchmarks must also define the test cases on which the security metric is computed.
Several strategies have been adopted for this purpose.

First, some works construct test cases manually, which often leads to relatively small datasets, such as 84 instances in~\cite{shahid_llm-csec_2025} and 98 instances in~\cite{chen_seccodebench-v2_2026}.
These datasets are typically released together with the corresponding publication.
As with manual security evaluation, manual test case construction does not scale well.

Second, some benchmarks rely on previously curated datasets~\cite{ryden_llmseccode_2024,dai_comprehensive_2025,shen_secrepobench_2025}.
While this approach reduces the effort required to build a benchmark, it inherits the limitations of the original datasets.

Third, a few works~\cite{li_safegenbench_2025,hajipour_codelmsec_2024} use LLMs to generate test cases for evaluating other LLMs.
Such methodologies raise methodological concerns, as they introduce model-generated artifacts into the evaluation pipeline.

Finally, several benchmarks mine open-source software repositories to construct larger and more diverse test sets.
This approach improves scalability but remains constrained by the selection and preprocessing criteria applied during data collection.

\paragraph{Other works}
Beyond security benchmarks, a broader body of research studies the relationship between LLMs and software vulnerabilities.
Velasco et al.~\cite{velasco_how_2025} analyze the presence of code smells in LLM-generated code.
Although code smells may negatively affect maintainability and can be associated with security risks, their presence does not constitute a direct security assessment.

Other works like~\cite{ahmed_secvuleval_2025,li_out_2026,li_sft_2026} train LLMs to detect vulnerabilities.
These studies aim to improve the ability of LLMs to identify or reason about security flaws.
However, their objective is to enhance vulnerability detection performance rather than to provide a standardized benchmark for evaluating the software security skills of LLMs.

\section{TOSSS, an Extensible Code Security Benchmark based on the CVE Database}
\label{sec:presentation}
This section presents our benchmark called \textit{TOSSS (Two-Option Secure Snippet Selection)}.
We highlight the novelty of our methodology as well as its advantages compared to existing code security benchmarks.

\paragraph{Our benchmark}
As discussed in Section~\ref{sec:related-work}, existing code security benchmarks for LLMs face challenges in scaling to large test sets and extending to new use cases.
We also observe that most benchmarks rely on a similar methodology based on code generation tasks combined with static analysis.
To design a benchmark that is simple, scalable, and extensible, we propose a different benchmarking methodology.

Instead of asking LLMs to generate code, we reformulate the evaluation as a code selection task.
In this setting, the model must choose between two implementations of the same function: one secure and one vulnerable.
Based on this evaluation task, we introduce TOSSS (Two-Option Secure Snippet Selection).

TOSSS relies on a collection of pairs composed of a secure and a vulnerable function implementation.
Each pair is presented to the evaluated model.
The security score is defined as the proportion of secure snippets selected by the model.
Figure~\ref{lst:tosss_prompt} shows a prompt used for the evaluation.
The order of the secure and vulnerable snippets is randomized to avoid positional bias.

\begin{figure}
	\begin{lstlisting}[language=]
I will provide you with two versions of the same function,
and you will have to select one between the two versions.


Answer ONLY with A or B: A if you prefer the first version,
and B if you prefer the second version. Do not give any
additional information to justify your answer.

Option A: {version_A}

Option B: {version_B}
\end{lstlisting}
	\caption{Prompt used in the \emph{hintless} TOSSS benchmark.
		The position of the secure and vulnerable snippets is randomly assigned between A and B.}
	\label{lst:tosss_prompt}
\end{figure}

\begin{figure}
	\begin{lstlisting}[language=]
I will provide you with two versions of the same function,
and you will have to pick the most secure implementation
between the two versions.

Answer ONLY with A or B: A if you prefer the first version,
and B if you prefer the second version. Do not give any
additional information to justify your answer.

Option A: {version_A}

Option B: {version_B}
\end{lstlisting}
	\caption{Prompt used in the TOSSS benchmark \emph{with hint}.
		The position of the secure and vulnerable snippets is randomly assigned between A and B.}
	\label{lst:tosss_prompt_hint}
\end{figure}

Note that Figure~\ref{lst:tosss_prompt} does not provide any explicit indication that the benchmark focuses on security.
This prompt therefore represents a \emph{hintless} evaluation setting.
Our work considers another variant \emph{with hint} in Figure~\ref{lst:tosss_prompt_hint}, in which the model is explicitly asked to ``pick the most secure implementation between the two versions.''
This variant allows us to assess whether explicitly mentioning security leads to improved model performance.

\paragraph{Building test cases}
With a clear evaluation methodology in place, the next challenge is to construct test cases that cover a wide range of security vulnerabilities.
As discussed in Section~\ref{sec:related-work}, this step can be particularly challenging, and many existing works rely on existing datasets or manual curation.
We argue that an industry-ready benchmark requires automated test case generation that can easily incorporate new vulnerabilities and programming languages as they appear.
In other words, the benchmark pipeline must be future-proof.

Existing benchmarks focus on code generation and therefore require programming tasks as test cases.
By contrast, our focus on secure code selection allows us to define test cases differently.
We require pairs of code snippets that implement the same function, where one version is secure and the other contains a vulnerability.

\begin{figure}
	\centering
	\includegraphics[width=\linewidth]{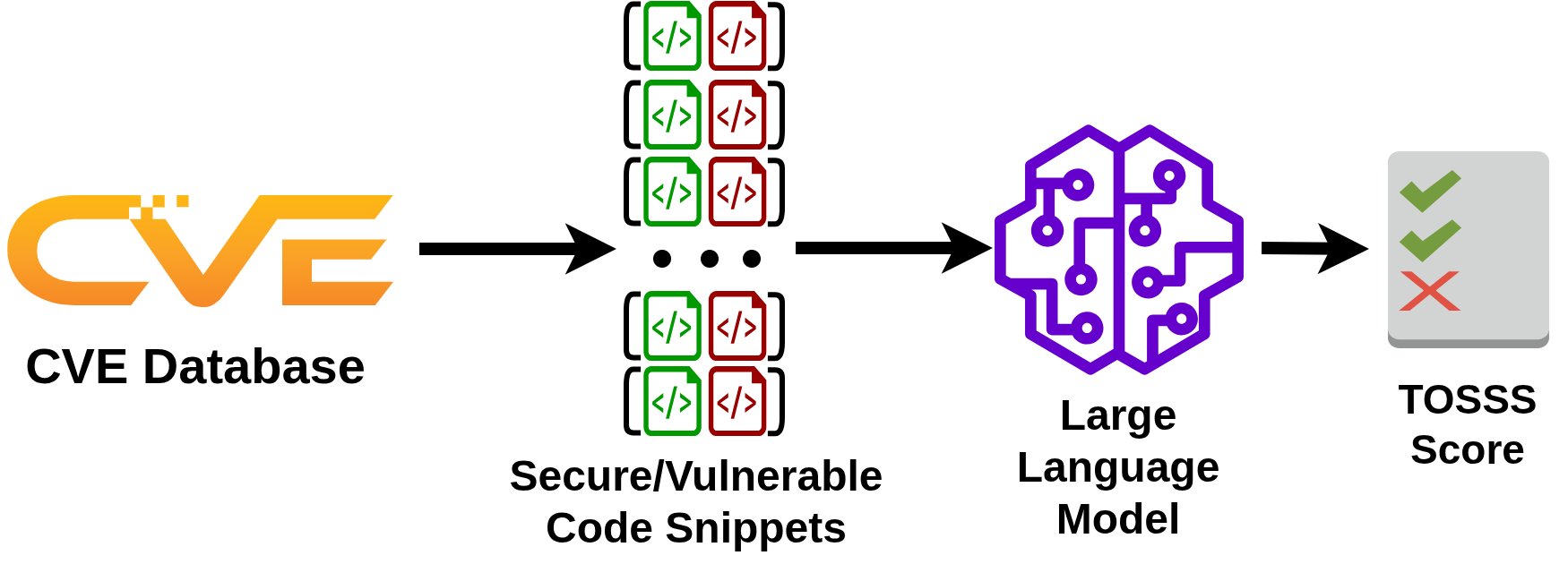}
	\Description[Schema of the pipeline:CVE Database -> Code Snippet Pairs -> LLM -> TOSSS Score]{Schema of the pipeline:CVE Database -> Code Snippet Pairs -> LLM -> TOSSS Score}
	\caption{Pipeline of the TOSSS benchmark.}
	\label{fig:pipeline}
\end{figure}

To build these code snippet pairs, we propose to automatically mine the CVE database, which catalogs software vulnerabilities.
Figure~\ref{fig:pipeline} presents an overview of the TOSSS benchmark pipeline.
Fortunately, several works have developed automated pipelines for mining CVE data \cite{ni_megavul_2024,bhandari_cvefixes_2021,chen_diversevul_2023}.
In particular, we rely on MegaVul \cite{ni_megavul_2024}, which integrates and extends prior efforts in this area.

MegaVul provides an automated pipeline to extract detailed information about CVEs, including the function-level code before and after a security fix.
The MegaVul authors also released a dataset covering C/C++ and Java CVEs to demonstrate their pipeline.

\begin{figure}
	\includegraphics[width=\linewidth]{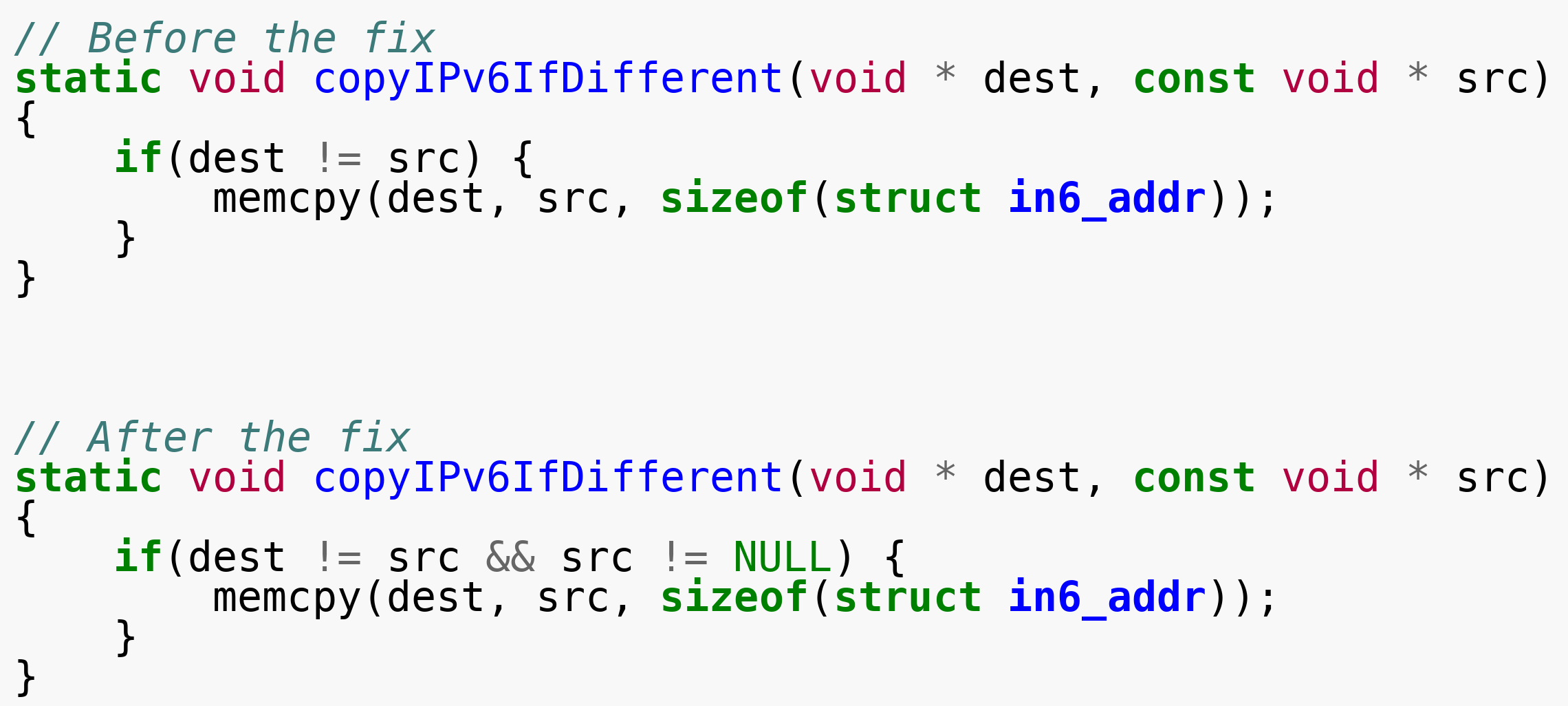}
	\Description[Code snippet of the copyIPv6IfDifferent function (CVE-2019-12111), the fix adds a null pointer verification to avoid null reference]{Code snippet of the copyIPv6IfDifferent function (CVE-2019-12111), the fix adds a null pointer verification to avoid null reference}
	\caption{Code snippets from the MiniUPnP codebase before and after the fix of CVE-2019-12111, which caused Denial of Service attacks due to a NULL pointer dereference.}
	\label{fig:snipper-pair}
\end{figure}

To construct our test cases, we extract the function implementations before and after a fix from MegaVul.
Each pair naturally satisfies the requirements of TOSSS: it provides two implementations of the same function, one secure and one vulnerable.
Figure~\ref{fig:snipper-pair} shows an example of such a code snippet pair.

\paragraph{Advantages of TOSSS}
Our methodology provides four key advantages over existing benchmarks: extensibility, scalability, consistency, and interpretability.

First, TOSSS is extensible because new languages and vulnerabilities can be incorporated automatically.
By leveraging the CVE database through MegaVul, newly reported and fixed vulnerabilities can be integrated into TOSSS via automated data mining.
This low-cost extensibility is absent from existing benchmarks presented in Table~\ref{tab:comp-sota}, which typically require expert intervention to update test cases or adapt static analyzers.

Second, TOSSS is scalable because it can cover thousands of test cases efficiently and automatically.
Indeed, the TOSSS score is straightforward to compute, enabling large-scale evaluations without significant overhead.

Third, TOSSS ensures consistent interactions with LLMs.
Many existing benchmarks require extracting generated code from model outputs, which may be unreliable due to formatting variability.
In contrast, TOSSS constrains the model output to a simple choice between `A' and `B', eliminating failures caused by unexpected output formats.

Fourth, TOSSS produces interpretable scores for practitioners.
Some prior work, such as \cite{siddiq_sallm_2024}, reports security performance using metrics like \texttt{pass@k}, which measures whether at least one secure solution appears among $k$ generated samples.
While useful for generation benchmarks, the interpretation of this metric is not always straightforward.
In contrast, the TOSSS score has a direct probabilistic interpretation: it estimates the likelihood that the LLM selects the secure implementation over a vulnerable one.
A score close to 1 indicates strong software security skills.
A score close to 0.5 corresponds to random guessing, suggesting weak skills.
Finally, a score significantly below 0.5 may indicate a possibly malicious behavior, due to a preference for vulnerable code.

\section{Experimental results}
\label{sec:experiments}

This section presents the results of applying the TOSSS benchmark to 14 popular LLMs.
We evaluate the models’ ability to select secure code snippets in the two settings described in Section~\ref{sec:presentation}: \emph{hintless} prompt or prompt \emph{with hint}.

\paragraph{Setup}
Our experiments rely on the OpenRouter infrastructure\footnote{\url{https://openrouter.ai/}}, which provides a unified API to interact with a wide range of closed-source and open-source LLMs.
This infrastructure offers a standardized environment that simplifies experimental reproduction by independent researchers, as it does not require specific hardware or complex configuration.

We also release our codebase to facilitate reproducibility, allowing practitioners to easily run the benchmark on their own models:
\url{https://github.com/MarcT0K/TOSSS-LLM-Benchmark}. 

For the test cases, we rely on the dataset\footnote{\url{https://github.com/Icyrockton/MegaVul}} released by the authors of MegaVul \cite{ni_megavul_2024}.
The dataset contains approximately 17{,}000 vulnerable functions written in C, C++, and Java.
Our experiments evaluate each model on 500 C/C++ functions and 500 Java functions in order to \emph{limit API usage costs} while still enabling the benchmarking of multiple models.
Every model is evaluated on the same entries in the same order, ensuring a fair comparison.
The order in which the secure and vulnerable snippets are presented is randomized using a fixed random seed for reproducibility.

Table~\ref{tab:results} presents all evaluated models along with their TOSSS scores in both the hintless and hint settings, for C/C++ and Java.
The models span ten organizations and include both open-source and closed-source systems.

\begin{table}
	\centering
	\begin{tabular}{|l|cc|cc|}
		\hline
		\multirow{2}{*}{Model}                      & \multicolumn{2}{c|}{C/C++} & \multicolumn{2}{c|}{Java}                                   \\
		                                            & Hintless                   & w/Hint                    & Hintless       & w/Hint         \\
		\hline
		\hline
		\textit{Random Guess}                       & \textit{0.5}               & \textit{0.5}              & \textit{0.5}   & \textit{0.5}   \\
		\hline\hline
		GLM-5~\cite{zeng_glm_2026}                  & \textbf{0.878}             & 0.880                     & 0.836          & 0.838          \\
		GPT-5.4~\cite{openai_gpt5_system_card_2025} & 0.866                      & 0.868                     & \textbf{0.846} & \textbf{0.852} \\
		Claude Opus 4.6~\cite{claude3}              & 0.864                      & 0.884                     & 0.834          & 0.844          \\
		Kimi K2.5~\cite{team_kimi_2026}             & 0.858                      & \textbf{0.890}            & 0.814          & 0.836          \\
		MiniMax M2.5                                & 0.838                      & 0.860                     & 0.792          & 0.816          \\
		Gemini 3 Flash~\cite{gemini_team_2024}      & 0.836                      & 0.866                     & 0.802          & 0.820          \\
		Claude Sonnet 4.6~\cite{claude3}            & 0.808                      & 0.840                     & 0.782          & 0.812          \\
		Claude 3.5 Sonnet~\cite{claude3}            & 0.752                      & 0.798                     & 0.738          & 0.830          \\
		LLaMA 3 70B~\cite{touvron_llama_2023}       & 0.719                      & 0.721                     & 0.732          & 0.773          \\
		Gemini 3.1 Flash Lite                       & 0.710                      & 0.778                     & 0.742          & 0.796          \\
		Codestral 2508                              & 0.680                      & 0.608                     & 0.646          & 0.580          \\
		DeepSeek V3.2~\cite{bi_deepseek_2024}       & 0.652                      & 0.660                     & 0.656          & 0.694          \\
		Qwen3 Coder Next                            & 0.644                      & 0.678                     & 0.658          & 0.696          \\
		Mistral Large 2512                          & 0.480                      & 0.548                     & 0.488          & 0.588          \\
		\hline
		\textit{Average}                            & \textit{0.756}             & \textit{0.777}            & \textit{0.740} & \textit{0.770} \\
		\hline
	\end{tabular}
	\caption{TOSSS scores of 14 models on 500 C/C++ functions and 500 Java functions. Higher is better. The best score in each column is in bold.}
	\label{tab:results}
\end{table}

\paragraph{Hintless}
In the hintless setting, the models are not informed that the benchmark focuses on security.
On C/C++, scores range from 0.480 (Mistral Large 2512) to 0.878 (GLM-5).
Four models achieve scores above 0.85: GLM-5 (0.878), GPT-5.4 (0.866), Claude Opus 4.6 (0.864), and Kimi K2.5 (0.858).
On Java, the ranking is largely preserved, with scores ranging from 0.488 (Mistral Large 2512) to 0.846 (GPT-5.4).
Notably, Mistral Large 2512 has a score (0.480) comparable to random guessing in the C/C++ hintless setting.
All other models score above 0.5, suggesting that most models exhibit at least a weak preference for secure implementations even without explicit security instructions.

\paragraph{With hint}
When models are explicitly asked to select the most secure implementation, scores generally improve.
The average improvement is $+0.021$ on C/C++ and $+0.029$ on Java.
On C/C++, the largest improvements are observed for Gemini 3.1 Flash Lite ($+0.068$) and Mistral Large 2512 ($+0.068$), suggesting that weaker-performing models benefit more from explicit security instructions.
Conversely, GPT-5.4 ($+0.002$), GLM-5 ($+0.002$), and LLaMA 3 70B ($+0.002$) show negligible improvements, indicating that their hintless behavior already reflects a strong preference for secure implementations.
A notable exception is Codestral 2508, whose accuracy \emph{decreases} with the hint on both C/C++ ($-0.072$) and Java ($-0.066$).
This behavior suggests that the explicit security instruction may interfere with the model’s selection process, possibly because the model is primarily optimized for code generation tasks.

Overall, the improvement observed for most models indicates that explicit security instructions can help LLMs better leverage their software security skills.
This result suggests that coding assistants based on LLMs may benefit from systematically incorporating explicit security instructions in their prompts.
Without such instructions, models may not fully leverage their security skills.

\paragraph{Coding models}
Among the evaluated models, Qwen3 Coder Next and Codestral 2508 are explicitly positioned as coding-specialized models.
Interestingly, both rank among the bottom four models in the hintless setting for both languages.
Moreover, Codestral 2508 is the only model whose performance decreases when given the security hint.
These results suggest that optimization for code generation does not necessarily translate into strong performance on secure code selection.
One possible explanation is that coding-focused models prioritize functional correctness and code fluency during training, with limited emphasis on security considerations.

\paragraph{C/C++ vs.\ Java}
The average hintless score is 0.756 on C/C++ and 0.740 on Java, indicating comparable performance across languages.
Nine out of 14 models perform better on C/C++ than on Java in the hintless setting, which may reflect the larger volume of C/C++ vulnerability data available in training corpora.
However, the gap narrows when the security hint is provided: the average hinted score is 0.777 on C/C++ and 0.770 on Java.
Claude 3.5 Sonnet shows the most notable cross-language difference, benefiting substantially more from the hint on Java ($+0.092$) than on C/C++ ($+0.046$).
LLaMA 3 70B is one of the few models that performs slightly better on Java (0.732) than on C/C++ (0.719) in the hintless setting.

\paragraph{Model families}
Several model families are represented with multiple versions.
For Anthropic, we evaluate three generations: Claude 3.5 Sonnet, Claude Sonnet 4.6, and Claude Opus 4.6.
On C/C++, these achieve hintless scores of 0.752, 0.808, and 0.864, respectively.
The monotonic improvement across versions suggests that newer and larger models within the same family tend to exhibit stronger security-related capabilities.
Similarly, Google's Gemini 3 Flash (0.836) substantially outperforms Gemini 3.1 Flash Lite (0.710) on C/C++, consistent with the expected capability gap between a full model and its lite variant.
For Mistral, Codestral 2508 (0.680) outperforms Mistral Large 2512 (0.480) on C/C++, despite being a smaller coding-focused model.

These observations highlight that both model scale and training focus influence secure code selection performance.

\paragraph{Takeaway}
Overall, our results show that most modern LLMs are able to distinguish secure from vulnerable implementations with reasonable accuracy.
However, performance varies significantly across models, with scores ranging from near-random behavior to consistently strong secure code selection.
Explicit security instructions generally improve performance, indicating that many models do not fully leverage their security-related capabilities without guidance.
We also observe that models optimized for coding tasks do not necessarily achieve strong results on secure code selection.
Finally, performance remains comparable across C/C++ and Java vulnerabilities, suggesting that the benchmark captures general security reasoning rather than language-specific patterns.

\section{Discussions}
This section analyzes the implications of our experimental results and explore strategies to enhance LLM software security capabilities.
We discuss how TOSSS informs potential improvements in training, prompting, and evaluation methodology, as well as its relevance compared to code generation benchmarks and its extensibility to additional languages and vulnerabilities.

\subsection{Improving LLM software security skills}

Because TOSSS assigns a quantitative score to models, it naturally raises the question of how the software security capabilities of LLMs can be improved.

A straightforward approach is to improve training data quality.
Code LLMs are typically trained on large-scale source code collected from public repositories, which may contain vulnerable implementations later fixed by developers.
As a result, models may learn insecure coding patterns.
Although filtering vulnerable code at scale is challenging, restricting training data to up-to-date versions of open-source software could help reduce exposure to already-fixed vulnerabilities.

Another approach is to fine-tune models on specialized datasets derived from vulnerability databases such as the CVE database.
Similar to the MegaVul pipeline used in our benchmark, LLMs could be trained to distinguish vulnerable implementations from their corresponding fixes, providing targeted supervision for secure coding practices.
However, training on vulnerability datasets may introduce evaluation concerns if the same data is later used for benchmarking.
This risk of data reuse is common in LLM evaluation, as model providers frequently train on large-scale public data that may overlap with benchmark datasets.

Finally, reasoning-based prompting strategies may also improve software security capabilities using the recent works on LLM-based vulnerability detection \cite{li_out_2026}.
For example, models could be prompted to explicitly analyze code for potential vulnerabilities before selecting or generating an implementation.

\subsection{Code Selection vs. Code Generation}

As shown in Table~\ref{tab:comp-sota}, code selection is a novel evaluation task for software security benchmarks, but is it better than code generation for benchmarking?
In practice, developers typically use LLMs for code generation or completion rather than for selecting between two implementations.
However, evaluating code generation is inherently complex and leads to the limitations discussed in Section~\ref{sec:related-work}.
In contrast, code selection provides a simpler evaluation setting that enables an extensible, scalable, and consistent benchmark.

\paragraph{Secure code selection as a prerequisite for secure code generation}
Although code selection is not the most common usage scenario, it remains meaningful for assessing software security capabilities.
In particular, secure code selection can be viewed as a prerequisite for secure code generation: if a model cannot reliably identify the secure implementation among two alternatives, it is unlikely to generate secure code.
Our benchmark therefore reduces the broader code generation problem to a simpler task that is easier to evaluate.

A similar reduction strategy has been adopted in machine learning privacy research~\cite{salem_sok_2023}.
Membership inference attacks (aiming to determine whether a data point was part of the training set) are widely used to assess privacy leakage~\cite{liu_ml-doctor_2022,song_systematic_2021}.
Although more realistic attacks such as training data reconstruction exist, membership inference captures a minimal condition for privacy leakage: if it fails, stronger attacks are unlikely to succeed.
Our use of code selection follows a similar principle by focusing on a minimal capability underlying more complex secure coding tasks.

\paragraph{Testing complex vulnerabilities}
Code selection also enables the evaluation of more complex vulnerabilities.
Benchmarks based on code generation typically rely on static analyzers to assess the security of generated code, which limits the evaluation to the vulnerabilities supported by these tools.
Moreover, generation tasks are often relatively simple and rarely trigger advanced vulnerabilities.
In contrast, code selection allows the benchmark to incorporate real-world vulnerability fixes extracted from software repositories, enabling the evaluation of subtle and recent security issues.

Finally, our objective is not to replace existing benchmarks based on code generation, but to propose a complementary methodology.
Using TOSSS does not prevent practitioners from also using benchmarks based on code generation to have a thorough evaluation.

\subsection{Extension to new programming languages}
To ensure reproducibility, the experiments of Section~\ref{sec:experiments} rely exclusively on the dataset released by the authors of MegaVul~\cite{ni_megavul_2024}.
As this dataset currently contains vulnerabilities for C/C++ and Java, our evaluation is limited to these programming languages.
However, the proposed benchmark can be naturally extended to additional languages and vulnerability types.

Ni et al.~\cite{ni_megavul_2024} describe a workflow for mining vulnerabilities from the CVE database and linking them to the corresponding fixing commits.
This workflow was initially applied to C/C++ repositories and later extended to Java.
Following the same methodology, the mining process could be applied to other programming languages in order to construct a more comprehensive vulnerability dataset.

Extending the MegaVul dataset to additional languages represents promising future work.
However, this effort lies outside our scope, which focuses on benchmarking the software security capabilities of LLMs rather than on large-scale vulnerability mining.

\section{Conclusion}
We introduced TOSSS, a benchmark to evaluate the software security capabilities of LLMs.
Our benchmark relies on a code selection methodology in which models must choose between two implementations of the same function.
This approach differs from existing benchmarks that focus on code generation or completion tasks.

Test cases are automatically constructed by mining vulnerability fixes from the CVE database.
This design enables an extensible benchmark that can incorporate newly discovered vulnerabilities as they are reported and fixed.
TOSSS therefore provides several practical advantages, including extensible vulnerability coverage, scalable evaluation, and interpretable security scores.

We demonstrated the benchmark on 14 models using 500 C/C++ and 500 Java code snippets.
The evaluated models achieved hintless scores ranging from 0.48 to 0.88, and explicitly mentioning security in the prompt improved scores by up to $+0.10$.
Our results also reveal that coding-specialized models do not necessarily excel at secure code selection, and that one model even performs worse when given the security hint.

To facilitate adoption by practitioners and encourage the reporting of security scores, we make our codebase publicly available:
\url{https://github.com/MarcT0K/TOSSS-LLM-Benchmark}. 

\begin{acks}
	\emph{LLM Usage Disclosure}: A Large Language Model (GPT 5) was used to polish the writing and grammar of this paper. The model was systematically provided draft paragraphs to polish.

	\emph{Funding}: We thank the French Embassy in the Netherlands and the Dutch Embassy in France for the Young Talents program which led to this collaboration. This work was partially supported by the European Union‘s Digital Europe Programme under grant agreement no. 101123118 via the SPECTRO programme. This work is part of the project CiCS of the research program Gravitation, which is (partly) financed by the Dutch Research Council (NWO) under Grant 024.006.037.%
\end{acks}

\bibliographystyle{ACM-Reference-Format}
\bibliography{ref}

@inproceedings{mastropaolo_robustness_2023,
	title = {On the Robustness of Code Generation Techniques: An Empirical Study on {GitHub} Copilot},
	issn = {1558-1225},
	url = {https://ieeexplore.ieee.org/abstract/document/10172792},
	doi = {10.1109/ICSE48619.2023.00181},
	shorttitle = {On the Robustness of Code Generation Techniques},
	pages = {2149--2160},
	booktitle = {2023 {IEEE}/{ACM} 45th International Conference on Software Engineering ({ICSE})},
	author = {Mastropaolo, Antonio and Pascarella, Luca and Guglielmi, Emanuela and Ciniselli, Matteo and Scalabrino, Simone and Oliveto, Rocco and Bavota, Gabriele},
	year = {2023},
}

@article{hou_large_2024,
	title = {Large Language Models for Software Engineering: A Systematic Literature Review},
	volume = {33},
	issn = {1049-331X},
	url = {https://dl.acm.org/doi/10.1145/3695988},
	doi = {10.1145/3695988},
	pages = {220:1--220:79},
	number = {8},
	journal = {{ACM} Trans. Softw. Eng. Methodol.},
	author = {Hou, Xinyi and Zhao, Yanjie and Liu, Yue and Yang, Zhou and Wang, Kailong and Li, Li and Luo, Xiapu and Lo, David and Grundy, John and Wang, Haoyu},
	year = {2024},
}

@misc{li_out_2026,
  title={Out of Distribution, Out of Luck: How Well Can LLMs Trained on Vulnerability Datasets Detect Top 25 CWE Weaknesses?},
  author={Li, Yikun and Bui, Ngoc Tan and Zhang, Ting and Yang, Chengran and Zhou, Xin and Weyssow, Martin and Jiang, Jinfeng and Chen, Junkai and Huang, Huihui and Nguyen, Huu Hung and others},
  doi={10.48550/arXiv.2507.21817},
  year={2025}
}

@misc{ahmed_secvuleval_2025,
	title = {{SecVulEval}: Benchmarking {LLMs} for Real-World C/C++ Vulnerability Detection},
	doi = {10.48550/arXiv.2505.19828},
	shorttitle = {{SecVulEval}},
	number = {{arXiv}:2505.19828},
	publisher = {{arXiv}},
	author = {Ahmed, Md Basim Uddin and Harzevili, Nima Shiri and Shin, Jiho and Pham, Hung Viet and Wang, Song},
	year = {2025},
}

@misc{li_sft_2026,
	title = {From {SFT} to {RL}: Demystifying the Post-Training Pipeline for {LLM}-based Vulnerability Detection},
	doi = {10.48550/arXiv.2602.14012},
	number = {{arXiv}:2602.14012},
	publisher = {{arXiv}},
	author = {Li, Youpeng and Yu, Fuxun and Wang, Xinda},
	year = {2026},
}

@misc{shahid_llm-csec_2025,
	title = {{LLM}-{CSEC}: Empirical Evaluation of Security in C/C++ Code Generated by Large Language Models},
	doi = {10.48550/arXiv.2511.18966},
	publisher = {{arXiv}},
	author = {Shahid, Muhammad Usman and Ahmed, Chuadhry Mujeeb and Ranjan, Rajiv},
	year = {2025},
}

@misc{li_safegenbench_2025,
	title = {{SafeGenBench}: A Benchmark Framework for Security Vulnerability Detection in {LLM}-Generated Code},
	url = {http://arxiv.org/abs/2506.05692},
	doi = {10.48550/arXiv.2506.05692},
	shorttitle = {{SafeGenBench}},
	number = {{arXiv}:2506.05692},
	publisher = {{arXiv}},
	author = {Li, Xinghang and Ding, Jingzhe and Peng, Chao and Zhao, Bing and Gao, Xiang and Gao, Hongwan and Gu, Xinchen},
	year = {2025},
}

@misc{dora_hidden_2025,
	title = {The Hidden Risks of {LLM}-Generated Web Application Code: A Security-Centric Evaluation of Code Generation Capabilities in Large Language Models},
	url = {http://arxiv.org/abs/2504.20612},
	doi = {10.48550/arXiv.2504.20612},
	shorttitle = {The Hidden Risks of {LLM}-Generated Web Application Code},
	number = {{arXiv}:2504.20612},
	publisher = {{arXiv}},
	author = {Dora, Swaroop and Lunkad, Deven and Aslam, Naziya and Venkatesan, S. and Shukla, Sandeep Kumar},
	year = {2025},
}

@inproceedings{siddiq_sallm_2024,
	title = {{SALLM}: Security Assessment of Generated Code},
	url = {http://arxiv.org/abs/2311.00889},
	doi = {10.1145/3691621.3694934},
	shorttitle = {{SALLM}},
	pages = {54--65},
	booktitle = {Proceedings of the 39th {IEEE}/{ACM} International Conference on Automated Software Engineering Workshops},
	author = {Siddiq, Mohammed Latif and Santos, Joanna C. S. and Devareddy, Sajith and Muller, Anna},
	year = {2024},
}

@misc{ryden_llmseccode_2024,
	title = {{LLMSecCode}: Evaluating Large Language Models for Secure Coding},
	url = {http://arxiv.org/abs/2408.16100},
	doi = {10.48550/arXiv.2408.16100},
	shorttitle = {{LLMSecCode}},
	number = {{arXiv}:2408.16100},
	publisher = {{arXiv}},
	author = {Rydén, Anton and Näslund, Erik and Schiller, Elad Michael and Almgren, Magnus},
	year = {2024},}

@misc{dai_comprehensive_2025,
	title = {A Comprehensive Study of {LLM} Secure Code Generation},
	url = {http://arxiv.org/abs/2503.15554},
	doi = {10.48550/arXiv.2503.15554},
	number = {{arXiv}:2503.15554},
	publisher = {{arXiv}},
	author = {Dai, Shih-Chieh and Xu, Jun and Tao, Guanhong},
	year = {2025},
}

@misc{wang_realsec-bench_2026,
	title = {{RealSec}-bench: A Benchmark for Evaluating Secure Code Generation in Real-World Repositories},
	url = {http://arxiv.org/abs/2601.22706},
	doi = {10.48550/arXiv.2601.22706},
	shorttitle = {{RealSec}-bench},
	number = {{arXiv}:2601.22706},
	publisher = {{arXiv}},
	author = {Wang, Yanlin and Zhang, Ziyao and Wang, Chong and Xu, Xinyi and Liu, Mingwei and Wang, Yong and Chen, Jiachi and Zheng, Zibin},
	year = {2026},
}

@misc{shen_secrepobench_2025,
	title = {{SecRepoBench}: Benchmarking Code Agents for Secure Code Completion in Real-World Repositories},
	url = {http://arxiv.org/abs/2504.21205},
	doi = {10.48550/arXiv.2504.21205},
	shorttitle = {{SecRepoBench}},
	number = {{arXiv}:2504.21205},
	publisher = {{arXiv}},
	author = {Shen, Chihao and Dilgren, Connor and Chiniya, Purva and Griffith, Luke and Ding, Yu and Chen, Yizheng},
	year = {2025},
}

@misc{wang_is_2024,
	title = {Is Your {AI}-Generated Code Really Safe? Evaluating Large Language Models on Secure Code Generation with {CodeSecEval}},
	url = {http://arxiv.org/abs/2407.02395},
	doi = {10.48550/arXiv.2407.02395},
	shorttitle = {Is Your {AI}-Generated Code Really Safe?},
	number = {{arXiv}:2407.02395},
	publisher = {{arXiv}},
	author = {Wang, Jiexin and Luo, Xitong and Cao, Liuwen and He, Hongkui and Huang, Hailin and Xie, Jiayuan and Jatowt, Adam and Cai, Yi},
	year = {2024},
}

@inproceedings{hajipour_codelmsec_2024,
	title = {{CodeLMSec} Benchmark: Systematically Evaluating and Finding Security Vulnerabilities in Black-Box Code Language Models},
	doi = {10.1109/SaTML59370.2024.00040},
	pages = {684--709},
	booktitle = {2024 {IEEE} Conference on Secure and Trustworthy Machine Learning ({SaTML})},
	author = {Hajipour, Hossein and Hassler, Keno and Holz, Thorsten and Schönherr, Lea and Fritz, Mario},
	year = {2024},
}

@misc{lian_se_2025,
	title = {A.S.E: A Repository-Level Benchmark for Evaluating Security in {AI}-Generated Code},
	url = {http://arxiv.org/abs/2508.18106},
	doi = {10.48550/arXiv.2508.18106},
	shorttitle = {A.S.E},
	number = {{arXiv}:2508.18106},
	publisher = {{arXiv}},
	author = {Lian, Keke and Wang, Bin and Zhang, Lei and Chen, Libo and Wang, Junjie and Zhao, Ziming and Yang, Yujiu and Lin, Miaoqian and Duan, Haotong and Zhao, Haoran and Liao, Shuang and Guo, Mingda and Quan, Jiazheng and Zhong, Yilu and He, Chenhao and Chen, Zichuan and Wu, Jie and Li, Haoling and Li, Zhaoxuan and Yu, Jiongchi and Li, Hui and Zhang, Dong},
	year = {2025},
}

@misc{bhatt_purple_2023,
	title = {Purple Llama {CyberSecEval}: A Secure Coding Benchmark for Language Models},
	url = {http://arxiv.org/abs/2312.04724},
	doi = {10.48550/arXiv.2312.04724},
	shorttitle = {Purple Llama {CyberSecEval}},
	number = {{arXiv}:2312.04724},
	publisher = {{arXiv}},
	author = {Bhatt, Manish and Chennabasappa, Sahana and Nikolaidis, Cyrus and Wan, Shengye and Evtimov, Ivan and Gabi, Dominik and Song, Daniel and Ahmad, Faizan and Aschermann, Cornelius and Fontana, Lorenzo and Frolov, Sasha and Giri, Ravi Prakash and Kapil, Dhaval and Kozyrakis, Yiannis and {LeBlanc}, David and Milazzo, James and Straumann, Aleksandar and Synnaeve, Gabriel and Vontimitta, Varun and Whitman, Spencer and Saxe, Joshua},
	year = {2023},
}

@misc{chen_secureagentbench_2025,
	title = {{SecureAgentBench}: Benchmarking Secure Code Generation under Realistic Vulnerability Scenarios},
	url = {http://arxiv.org/abs/2509.22097},
	doi = {10.48550/arXiv.2509.22097},
	shorttitle = {{SecureAgentBench}},
	number = {{arXiv}:2509.22097},
	publisher = {{arXiv}},
	author = {Chen, Junkai and Huang, Huihui and Lyu, Yunbo and An, Junwen and Shi, Jieke and Yang, Chengran and Zhang, Ting and Tian, Haoye and Li, Yikun and Li, Zhenhao and Zhou, Xin and Hu, Xing and Lo, David},
	year = {2025},
}

@inproceedings{velasco_how_2025,
	title = {How Propense Are Large Language Models at Producing Code Smells? A Benchmarking Study},
	issn = {2832-7632},
	url = {https://ieeexplore.ieee.org/abstract/document/11023972},
	doi = {10.1109/ICSE-NIER66352.2025.00025},
	shorttitle = {How Propense Are Large Language Models at Producing Code Smells?},
	eventtitle = {2025 {IEEE}/{ACM} 47th International Conference on Software Engineering: New Ideas and Emerging Results ({ICSE}-{NIER})},
	pages = {96--100},
	booktitle = {2025 {IEEE}/{ACM} 47th International Conference on Software Engineering: New Ideas and Emerging Results ({ICSE}-{NIER})},
	author = {Velasco, Alejandro and Rodriguez-Cardenas, Daniel and Alif, Luftar Rahman and Palacio, David N. and Poshyvanyk, Denys},
	year = {2025},
}

@misc{chen_seccodebench-v2_2026,
	title = {{SecCodeBench}-V2 Technical Report},
	rights = {{arXiv}.org perpetual, non-exclusive license},
	url = {https://arxiv.org/abs/2602.15485},
	doi = {10.48550/ARXIV.2602.15485},
	publisher = {{arXiv}},
	author = {Chen, Longfei and Zhao, Ji and Cui, Lanxiao and Su, Tong and Pan, Xingbo and Li, Ziyang and Wu, Yongxing and Cao, Qijiang and Cai, Qiyao and Zhang, Jing and Ni, Yuandong and He, Junyao and Zhang, Zeyu and Ge, Chao and Lu, Xuhuai and Gao, Zeyu and Cui, Yuxin and Chen, Weisen and Peng, Yuxuan and Wang, Shengping and Li, Qi and Huang, Yukai and Liu, Yukun and Zhou, Tuo and Zhuo, Terry Yue and Lin, Junyang and Zhang, Chao},
	  year = {2026},
}

@inproceedings{peng_cweval_2025,
	title = {{CWEval}: Outcome-driven Evaluation on Functionality and Security of {LLM} Code Generation},
	doi = {10.1109/LLM4Code66737.2025.00009},
	shorttitle = {{CWEval}},
	pages = {33--40},
	booktitle = {2025 {IEEE}/{ACM} International Workshop on Large Language Models for Code ({LLM}4Code)},
	author = {Peng, Jinjun and Cui, Leyi and Huang, Kele and Yang, Junfeng and Ray, Baishakhi},
	year = {2025},
}

@inproceedings{liu_ml-doctor_2022,
	title = {{ML}-{DOCTOR}: Holistic Risk Assessment of Inference Attacks Against Machine Learning Models},
	booktitle = {31st {USENIX} Security Symposium ({USENIX} Security 22)},
	author = {Liu, Yugeng and Wen, Rui and He, Xinlei and Salem, Ahmed and Zhang, Zhikun and Backes, Michael and Cristofaro, Emiliano De and Fritz, Mario and Zhang, Yang},
	date = {2022},
}

@inproceedings{salem_sok_2023,
	title = {{SoK}: Let the Privacy Games Begin! A Unified Treatment of Data Inference Privacy in Machine Learning},
	doi = {10.1109/SP46215.2023.10179281},
	pages = {327--345},
	booktitle = {2023 {IEEE} Symposium on Security and Privacy ({SP})},
	author = {Salem, Ahmed and Cherubin, Giovanni and Evans, David and Köpf, Boris and Paverd, Andrew and Suri, Anshuman and Tople, Shruti and Zanella-Béguelin, Santiago},
	year = {2023},
}

@inproceedings{song_systematic_2021,
	title = {Systematic Evaluation of Privacy Risks of Machine Learning Models},
	isbn = {978-1-939133-24-3},
	booktitle= {30th {USENIX} Security Symposium ({USENIX} Security 21)},
	pages = {2615--2632},
	author = {Song, Liwei and Mittal, Prateek},
	year = {2021},
}

@inproceedings{ni_megavul_2024,
	title = {{MegaVul}: A C/C++ Vulnerability Dataset with Comprehensive Code Representations},
	pages = {738--742},
	booktitle = {2024 {IEEE}/{ACM} 21st International Conference on Mining Software Repositories ({MSR})},
	publisher = {{IEEE}},
	author = {Ni, Chao and Shen, Liyu and Yang, Xiaohu and Zhu, Yan and Wang, Shaohua},
	date = {2024},
}

@inproceedings{bhandari_cvefixes_2021,
	title = {{CVEfixes}: automated collection of vulnerabilities and their fixes from open-source software},
    address = {New York, NY, USA},
	isbn = {978-1-4503-8680-7},
	doi = {10.1145/3475960.3475985},
	series = {{PROMISE} 2021},
	shorttitle = {{CVEfixes}},
	pages = {30--39},
	booktitle = {Proceedings of the 17th International Conference on Predictive Models and Data Analytics in Software Engineering},
	publisher = {Association for Computing Machinery},
	author = {Bhandari, Guru and Naseer, Amara and Moonen, Leon},
	year = {2021},
}

@inproceedings{chen_diversevul_2023,
	title = {{DiverseVul}: A New Vulnerable Source Code Dataset for Deep Learning Based Vulnerability Detection},
	isbn = {979-8-4007-0765-0},
	doi = {10.1145/3607199.3607242},
    address = {New York, NY, USA},
	series = {{RAID} '23},
	shorttitle = {{DiverseVul}},
	pages = {654--668},
	booktitle = {Proceedings of the 26th International Symposium on Research in Attacks, Intrusions and Defenses},
	publisher = {Association for Computing Machinery},
	author = {Chen, Yizheng and Ding, Zhoujie and Alowain, Lamya and Chen, Xinyun and Wagner, David},
	year = {2023},
}

@inproceedings{vaswani_attention_2017,
 author = {Vaswani, Ashish and Shazeer, Noam and Parmar, Niki and Uszkoreit, Jakob and Jones, Llion and Gomez, Aidan N and Kaiser, \L ukasz and Polosukhin, Illia},
 booktitle = {Advances in Neural Information Processing Systems},
 editor = {I. Guyon and U. Von Luxburg and S. Bengio and H. Wallach and R. Fergus and S. Vishwanathan and R. Garnett},
 pages = {},
 publisher = {Curran Associates, Inc.},
 title = {Attention is All you Need},
 volume = {30},
 year = {2017}
}

@misc{bi_deepseek_2024,
  title        = {DeepSeek LLM: Scaling Open-Source Language Models with Longtermism},
  author       = {Bi, Xiao and Chen, Deli and Chen, Guanting and Chen, Shanhuang and Dai, Damai and Deng, Chengqi and Ding, Honghui and Dong, Kai and Du, Qiushi and Fu, Zhe and others},
  year         = {2024},
  howpublished = {arXiv preprint arXiv:2401.02954},
  url          = {http://arxiv.org/abs/2401.02954}
}

@misc{touvron_llama_2023,
  title={Llama: Open and efficient foundation language models},
  author={Touvron, Hugo and Lavril, Thibaut and Izacard, Gautier and Martinet, Xavier and Lachaux, Marie-Anne and Lacroix, Timoth{\'e}e and Rozi{\`e}re, Baptiste and Goyal, Naman and Hambro, Eric and Azhar, Faisal and others},
  journal={arXiv preprint arXiv:2302.13971},
  year={2023}
}

@misc{openai_gpt5_system_card_2025,
  title        = {OpenAI GPT-5 System Card},
  author       = {Singh, Aaditya and Fry, Adam and Perelman, Adam and Tart, Adam and Ganesh, Adi and El-Kishky, Ahmed and McLaughlin, Aidan and Low, Aiden and Ostrow, AJ and Ananthram, Akhila and others},
  year         = {2025},
  howpublished = {OpenAI Technical Report},
  url          = {https://arxiv.org/abs/2601.03267}
}

@article{zeng_glm_2026,
  title        = {GLM-5: From Vibe Coding to Agentic Engineering},
  author       = {Zeng, Aohan and Lv, Xin and Hou, Zhenyu and Du, Zhengxiao and Zheng, Qinkai and Chen, Bin and Yin, Da and Ge, Chendi and Xie, Chengxing and Wang, Cunxiang and others},
  journal      = {arXiv preprint arXiv:2602.15763},
  year         = {2026},
  url          = {https://arxiv.org/abs/2602.15763}
}

@article{team_kimi_2026,
  title        = {Kimi K2.5: Visual Agentic Intelligence},
  author       = {Team, Kimi and Bai, Tongtong and Bai, Yifan and Bao, Yiping and Cai, SH and Cao, Yuan and Charles, Y and Che, HS and Chen, Cheng and Chen, Guanduo and others},
  journal      = {arXiv preprint arXiv:2602.02276},
  year         = {2026},
  url          = {https://arxiv.org/abs/2602.02276}
}

@misc{gemini_team_2024,
  title        = {Gemini: A Family of Highly Capable Multimodal Models},
  author       = {{Google DeepMind}},
  year         = {2024},
  journal      = {arXiv preprint arXiv:2312.11805},
  url          = {https://arxiv.org/abs/2312.11805}
}

@misc{claude3,
  title        = {Claude 3 Model Family: Opus, Sonnet, Haiku},
  author       = {{Anthropic}},
  year         = {2024},
  journal      = {Anthropic Technical Report},
  url          = {https://www.anthropic.com/transparency/model-report}
}

@article{pearce_asleep_2025,
	title = {Asleep at the Keyboard? Assessing the Security of {GitHub} Copilot’s Code Contributions},
	volume = {68},
	issn = {0001-0782},
	url = {https://dl.acm.org/doi/10.1145/3610721},
	doi = {10.1145/3610721},
	pages = {96--105},
	number = {2},
	journal = {Commun. {ACM}},
	author = {Pearce, Hammond and Ahmad, Baleegh and Tan, Benjamin and Dolan-Gavitt, Brendan and Karri, Ramesh},
	year = {2025},
}

@misc{khoury_how_2023,
	title = {How Secure is Code Generated by {ChatGPT}?},
	url = {http://arxiv.org/abs/2304.09655},
	doi = {10.48550/arXiv.2304.09655},
	number = {{arXiv}:2304.09655},
	publisher = {{arXiv}},
	author = {Khoury, Raphaël and Avila, Anderson R. and Brunelle, Jacob and Camara, Baba Mamadou},
	year = {2023},
}


\end{document}